%% file: faba.tex
\documentclass[a4paper]{article}
\pdfoutput=1

\usepackage{dblfloatfix}
\usepackage{graphicx}
\usepackage{caption}  
\usepackage{subcaption}  
\usepackage{tikz}
\usetikzlibrary{shapes,arrows,calc,positioning}
\usepackage{amssymb}
\usepackage{amsmath}
\usepackage{paralist}
\usepackage{listings}
\usepackage[hyphens]{url}
\usepackage{csquotes}

\title{Entropy Non-increasing Games for the Improvement of Dataflow Programming}
\author{Norbert B\'atfai, Ren\'at\'o Besenczi, Gerg{\H o} Bogacsovics,\\Fanny Monori\thanks{All authors are with the Department of Information Technology, University of Debrecen, Hungary, \{batfai.norbert, besenczi.renato\}@inf.unideb.hu}}

\begin{document}
\maketitle

\input{abstract}

\input{introduction}

\input{facebattle}

\section{Entropy Non-increasing Interfaces of Games}

\input{entropy}

\input{conclusion}

\input{ack}

\bibliographystyle{alpha}
\bibliography{faba}

\end{document}

%% file: abstract.tex
\begin{abstract}
In this article, we introduce a new conception of a family of esport games called Samu Entropy to try to improve dataflow program graphs like the ones that are based on Google's TensorFlow. Currently, the Samu Entropy project specifies only requirements for new esport games to be developed with particular attention to the investigation of the relationship between esport and artificial intelligence. It is quite obvious that there is a very close and natural relationship between esport games and artificial intelligence. Furthermore, the project Samu Entropy focuses not only on using artificial intelligence, but on creating AI in a new way. We present a reference game called Face Battle that implements the Samu Entropy requirements.
\end{abstract}

\noindent\textbf{Keywords:} \textit{esport}, \textit{TensorFlow}, \textit{computation graphs}, \textit{gamification}, \textit{robopsychology}.

%% file: introduction.tex
\section{Introduction}
By now, playing massively multiplayer online (MMO) games has become a part of daily routine. Many computer game competitions and a whole new industry called esport is being built on the MMO game concept. The development of esports has been living in close symbiosis with the evolution of artificial intelligence research. The collaboration between Google DeepMind and Blizzard Entertainment\footnote{DeepMind and Blizzard to release StarCraft II as an AI research environment, \url{https://deepmind.com/blog/deepmind-and-blizzard-release-starcraft-ii-ai-research-environment/}} can also be interpreted as a next milestone in the development of artificial intelligence. In the present paper, we are going to introduce a new esport game family concept called Samu Entropy (or ESAMU, for short) \cite{ESAMUGitHub}, \cite{ESAMUInfTars}. The ESAMU concept is outlined in the English translation of its developer's guide\footnote{Pre-release of the Samu Entropy documentation, \url{https://github.com/nbatfai/SamuEntropy/releases/tag/v0.0.1}.} \cite{ESAMUDevGuide}.

Automated development of artificial convolutional neural network architectures is a hot topic in machine learning (ML). 
For example, papers \cite{TFArc} and \cite{CNNArc} present reinforcement learning approaches to develop the computing models, where the former one is based on TensorFlow (TF) computational graphs \cite{TF} and the latter one uses Q-learning. The authors of paper \cite{EvolArc} reported an evolutionary computing approach. As the main result of this paper, we introduce a new esport based approach to refine TensorFlow computational graphs. Specifically, we introduce a concept of an esport game with which a player will be able to search better neural network architectures. 

To embed a scientific challenge into a computer game is a well-known idea. The examples for this have ranged from science fiction stories to hard-core science (see for example the science fiction book SGU \cite[pp. 32]{SGU} or the scientific puzzle game called \textit{Foldit} \cite{Foldit}). It is important to note that ESAMU is not a typical human-computing \cite{hcgames} or citizen science \cite{csgames} game but, in contrast with the previously mentioned games, it is intended to become a killer application. At the present time, applications based on machine learning are becoming closer and closer to the normal everyday life. Nonetheless, developing deep learning applications is still a scientific task which requires highly trained and experienced professionals from this field. In our vision, an application, which implements the ESAMU concepts, can take the whole machine learning science closer to the public by giving a simplified interface for AI algorithms and methods. Certainly, killer apps cannot be \enquote{developed}, as nobody can decide whether a computer program becomes successful or not. But, with developing such an application we hope that it will gain a high level of attention, not only from the scientific community, but from a large segment of the public. Accordingly, ESAMU is not a specific game but only a specification set for the games to be developed.

It seems that TensorFlow may be the Pax Romana of machine learning programming (paper \cite{TFTour} shows a timeline of machine learning programming languages and the second one of the milestone works \cite{DMNature1} and \cite{DMNature2} used TensorFlow\footnote{See \url{https://research.googleblog.com/2016/01/alphago-mastering-ancient-game-of-go.html}}). TensorFlow is an open source heterogeneous machine learning software platform for running TensorFlow computational graphs on CPU, GPU, TPU, Android or iOS. By using computational graphs in a central role, TensorFlow (Python or C++ based) API programming can be regarded as a kind of classical dataflow programming \cite{DFP}, \cite{Kahn}.

In the following, a short description of the ESAMU concept \cite{ESAMUDevGuide} will be presented by way of a specific game called Face Battle. We focus on the question of embedding a TensorFlow computational graph into a computer game. At the current stage of the research, the graph of MNIST\footnote{MNIST For ML Beginners, \url{https://www.tensorflow.org/tutorials/mnist/beginners/}} tutorial example \cite{TFGitHub} will be used.

%% file: facebattle.tex
\section{Face Battle}

In the ESAMU esport family, every game will be developed following five arche\-types, so all ESAMU games must implement these behaviors. Every game will implement some sort of machine learning task, so we require a dedicated archetype for implementing AI methods. This first archetype is called \enquote{Samu, the Brain}. Practically, this archetype uses currently available machine learning algorithms and paradigms (with the primary focus on deep learning), and it can be extended in the future. Because the proper usage and sometimes even the understanding of these machine learning primitives can be difficult for non-experts, we use the \enquote{Gr{\'e}ta, the Builder} (GTB) archetype for fine-tuning end-to-end machine learning \enquote{flows}. In our vision, one implementation of this archetype could be an application where the user can fine tune the dataflow graph of Inception v3 \cite{Inception3} if the player wants to create an application suitable for object recognition. The apps implementing Samu and Gr\'eta are universal. That means that these two archetypes are independent from a specific implementation of ESAMU, and can be used in other ESAMU games. The apps from the archetype of Samu execute TF graphs and the apps from the archetype of Gr\'eta give the possibility to edit and fine-tune that TF graph. The \enquote{N{\'a}ndi, the Teacher} archetype is applied for the implementation of software that can use supervised learning. The archetype called \enquote{Matyi, the Hunter} consists of software that provide perception and intervention, e.g. in the Face Battle application, collecting images of faces can be considered as a software for this archetype. The archetype \enquote{Erika, the Fighter} realizes the competition aspect of ESAMU games, e.g. in the Face Battle game, comparing the accuracy of face recognition can be considered as a simple implementation of this archetype, but its main motivation is to provide an esport experience for ESAMU games (i.e. esport competitions that can be organized in an arena).

In the rest of this section, we give a description about how we use these archetypes in the Face Battle game.

\input{samubrain}

\input{motherboard}

\subsection{N\'andi, the Teacher}

This archetype provides supervised learning in games where it is necessary. In the game Face Battle, this archetype will implement a set of possible annotations and corrections related to the images taken by players. This will allow players to connect persons, emotions, locations, etc. to images, and to correct any possible misclassification (e.g. emotion recognition).

\subsection{Matyi, the Hunter}

This archetype is responsible for perception, data collection and, in some cases, intervention. In Face Battle, it will provide photo acquisition. It is important to implement this archetype in Face Battle, because we want to collect images about faces only. Therefore, some sort of pre-processing is required, e.g. face segmentation or alignment. In addition, this part of the software will have a function to recognize emotions too.

\subsubsection{Face Battle Dataset}

For every ESAMU game, distinct datasets are required to perform machine learning tasks. Using open datasets for testing face recognition methods are becoming a common practice, the number of open datasets is growing (see e.g.: \cite{LFW}, \cite{yi2014learning}, \cite{liu2015faceattributes}, \cite{6140979}, \cite{7780896}, \cite{7025068}). In our basic concept, every game has its own dataset and every dataset is open, moreover, every subset is linked to a certain player, who collected the corresponding data, via a social network profile. To the best of our knowledge, this will be the first public dataset where images are linked to people, except for datasets related to celebrities.

\subsection{Erika, the Fighter}

This archetype will offer the possibility of building a competition around Face Battle. This will be the part of the game where the players can compare each other's \enquote{learning architecture}, or more precisely, the efficiency (i.e. the error rate) of their face classification procedure.

One use could be the following: players will give each other some pictures about themselves. After learning these pictures, the accuracy of classification of other pictures of the same player will be compared. The player with the better accuracy wins the battle.

%% file: samubrain.tex
\subsection{Samu, the Brain}

The main purpose of this archetype is to execute TF graphs. Our main question is, what is the most suitable ML architecture for a corresponding task?

There are several well-known ML architectures that perform well on various datasets. In the past few years, convolutional neural networks (CNN) \cite{LeCun98} have made a reasonable break-through in the field. One of the earliest works that had a great impact was reported by Krizhevsky et al. in \cite{ImageNet}. This architecture used five convolutional layers and three fully connected layers and introduced several new solutions (e.g. \enquote{dropout}). Another approach is the Inception architecture that is first introduced in \cite{GoogLeNet}. This architecture was 22 layers deep, which ignited the spread of very deep CNNs. Later, the Inception architecture was improved to achieve better performance \cite{Inception3}. The aforementioned solutions were trained and tested on the ImageNet open dataset. Face recognition is also considered as a classical AI task. One example is DeepFace \cite{DeepFace} and its improved version FaceNet \cite{FaceNet}, both were trained and tested on Labeled Faces in the Wild and YouTube Faces dataset, and set the state-of-the-art performance in its time of publishing. Since the dataflow graph of these architectures are huge (tens of thousands of nodes in some cases), we are planning to provide a basic version that the player can fine-tune, so they do not need to build it from scratch. In the game Face Battle, this could be the dataflow graph of FaceNet.

These examples show us that some sort of a natural evolution can be seen in machine intelligence. This evolution is obviously powered by science and the natural need to create more and more efficient AI algorithms. But, this creation requires highly trained and experienced professionals. With the ESAMU concept, we try to create an architecture that can facilitate this evolution by connecting non-expert gamers to the AI research communities.

%% file: motherboard.tex
\subsection{Gr\'eta, the Builder}
\label{motherboard}

\begin{figure}
	\centering\includegraphics[scale=.35]{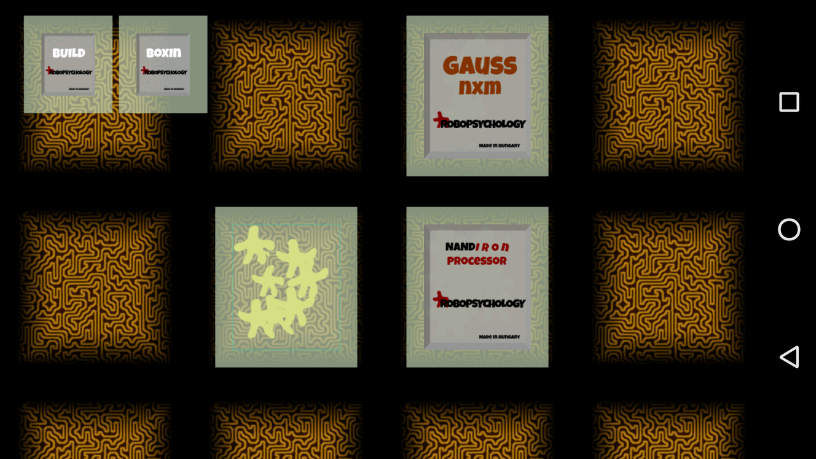}
	\caption{A screenshot of the app Motherboard Builder, which is a prototype for the GTB archetype.\label{fig_samubrainboard}}
\end{figure}

In the game Face Battle, the Motherboard Builder is our first rapid prototype\footnote{See in repository \url{https://github.com/nbatfai/SamuEntropy/tree/master/FACE_BATTLE/Greta/GretaTheBuilder}} for the GTB archetype. One key point of this motherboard approach is that we can place various objects related to computing on its surface. The idea came from the appearance and functionality of a motherboard of a desktop computer. Just like a real one, it can be equipped with several different \enquote{devices}, which can be changed, modified or upgraded later. In the aspect of an ESAMU game, these objects can be TF nodes (e.g. a CNN layer). It is not rare for a TensorFlow graph to have several thousand nodes. For example, the model of the TF tutorial example MNIST contains 137 nodes, but more sophisticated ones can contain tens of thousands of nodes. Building this from scratch on a screen or reviewing such a huge graph is impossible. Our aim is to simplify the visualization of complex ML architectures, so it would be easier to view the whole graph and fine-tune individual nodes. One of our main research question is: how can a complex TensorFlow graph be visualized so that the possibility to fine-tune individual nodes could remain. In section \ref{gtb}, we will present some more ideas about the GTB archetype. Figure \ref{fig_samubrainboard} shows a screenshot of Motherboard Builder.

%% file: entropy.tex
\begin{figure}[t!]
\centering
\begin{subfigure}[b]{\textwidth}
	\centering\includegraphics[scale=.4]{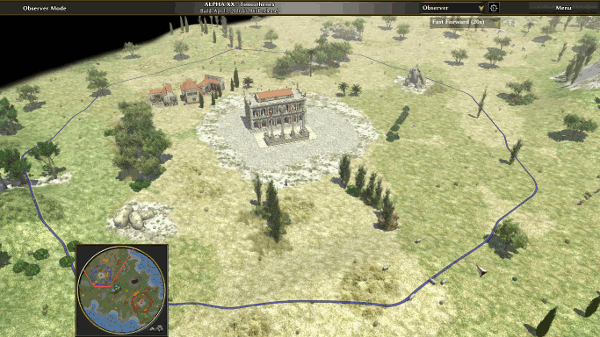}\label{01_ad}
\end{subfigure}
\par\bigskip
\begin{subfigure}[b]{\textwidth}
	\centering\includegraphics[scale=.4]{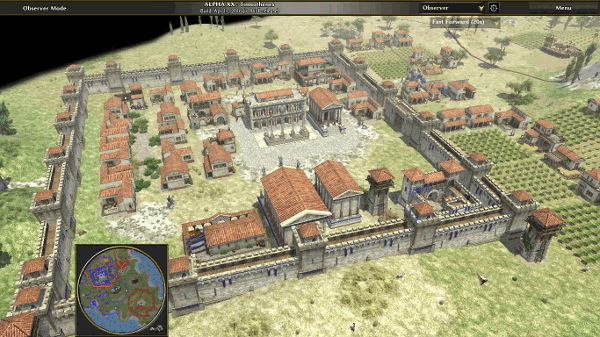}\label{32_ad}
\end{subfigure}
\caption{An illustration of \enquote{entropy} non-increasing property with two moments in the game \textit{0 A.D.}\label{fig_ad}}
\end{figure}

First of all, it should be noted that the entropy non-increasing property of elements of the game interface should be understood only as an intuitive metaphor, because the entropy of game's display is nearly independent from the order of what is viewed in the display (see \cite[pp. 402]{PenroseEmperor} for an analogous example of a similar situation). The inspiration behind the idea of the game having an entropy non-increasing property is RTS (Real-Time Strategy) games we play. Exemplary well-known RTS games are 
\textit{Age of Empires}\footnote{\url{https://www.ageofempires.com/}}, 
\textit{Warcraft}\footnote{\url{https://worldofwarcraft.com}}, 
\textit{StarCraft}\footnote{\url{http://www.battle.net/sc2}}, or the 
\textit{Cossacks}\footnote{\url{http://gsc-game.com/}} series, which were among the first generation of RTS games. Newer games like 
\textit{Clash of Clans}\footnote{\url{http://supercell.com/en/games/clashofclans/}}, or the open-source 
\textit{0 A.D.}\footnote{\url{https://play0ad.com/}, \url{https://github.com/0ad/0ad}} also have great player base.
The environment in these games may differ somewhat, but the main tasks a player can perform are roughly the same. In an RTS game, the players usually start with an empty map without buildings and full of resources. Then by collecting these resources, the player can build more buildings, create playable units, improve their structures and generally move forward in the game.
While playing with RTS games one can observe that the more the game progresses the more arranged it is. We believe that the game has its \enquote{entropy} at the highest in the beginning of the game with the empty map. With every \enquote{good} move the player makes the more ordered it becomes, so in our interpretation its \enquote{entropy} is decreasing. By good move, we intuitively mean a constructive event made by the player that moves him closer to winning.
In figure \ref{fig_ad} we illustrate this idea with two moments from the \textit{0 A.D.} game. We think in some cases this non-increasing property can also be perceived visually, as the map of the game looks more arranged as the game progresses.
The part of the game that implements the GTB archetype will have functions that are somewhat similar to building and creating activities in RTS games.

\input{gretabuilder}

\input{infacc}

%% file: gretabuilder.tex
\subsection{Towards the Gamification of GTB}
\label{gtb}

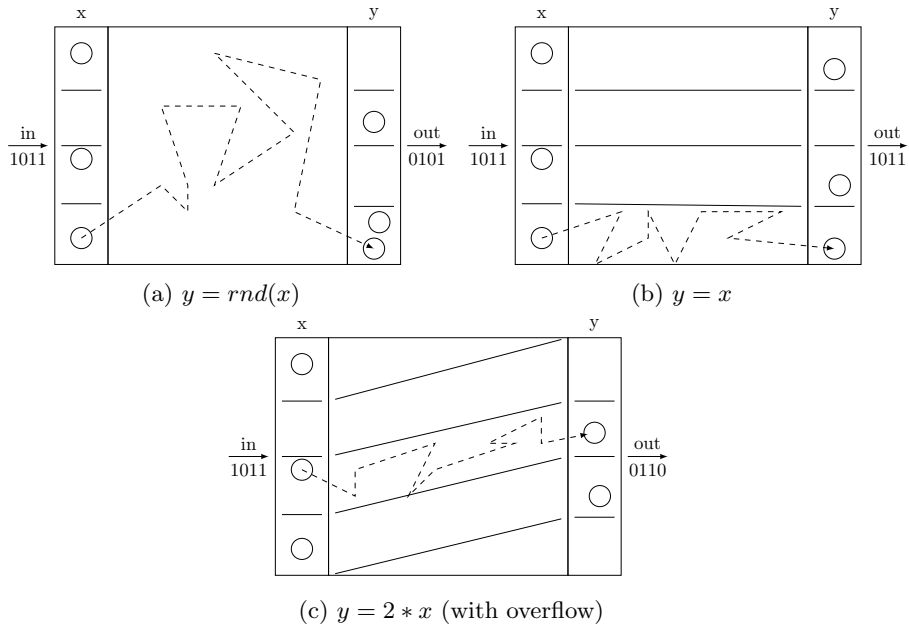
\begin{figure*}[t!]
\centering
\begin{subfigure}[t]{0.49\textwidth}
	\centering\scalebox{.7}{\input{fbox1}}\caption{$y=rnd(x)$\label{fig_fbox1}}
\end{subfigure}
\begin{subfigure}[t]{0.49\textwidth}
	\centering\scalebox{.7}{\input{fbox2}}\caption{$y=x$\label{fig_fbox2}}
\end{subfigure}
\begin{subfigure}[t]{\textwidth}
	\centering\scalebox{.7}{\input{fbox3}}\caption{$y=2*x$ (with overflow)\label{fig_fbox3}}
\end{subfigure}
\caption{This figure shows a fictional computing device where the programming is done by arrangement of the box. The first box (an idealization of a computer program) has no internal walls. The second one contains 3 internal walls and the third one contains 4 internal walls. Accordingly, it follows that the other two programs have smaller entropy than the first one.\label{fig_fboxs}}
\end{figure*}

The main purpose of the GTB archetype is to answer the research question: how can a TensorFlow computational graph be converted into the interface of a massively multiplayer game. At the current stage of the research, it seems that it is reasonable to focus on some MMORTS (Massively Multiplayer Online Real-Time Strategy) game to be developed from scratch.

In a typical RTS game, the elements of the game (e.g. buildings, army, town hall and other items) will become more and more sophisticated (or more complex) as times goes on. The ESAMU specification suggests measuring the development of these items in bits. From our point of view, the user interface of the game to be developed is a different visualization of a corresponding TensorFlow computation graph. After all, we would like to measure the goodness (the level of development) of elements of a computation graph. So, we would like to measure the goodness of the source code of pieces of the graph. First, we focus on the development of the whole source code as execution time goes on\footnote{In our present approach, the changing neural network weights are considered as part of the source code.}.

Let us start with the question, how can we measure the order of a source code? For example, how can we measure the order of a C or Python source code? There are several classical complexity measures, for example the cyclomatic complexity \cite{CyclomaticComplexity}, but these can measure only some features of the source code. The cyclomatic complexity measures how readable the investigated source code snippet is but it cannot tell us about the real goodness of the code. That is why these measures cannot be considered objective. For a totally objective measure we could use some Kolmogorov complexity based measure, but it would not be computable due to the fact that the Kolmogorov complexity is not recursive (here it should be noticed that the similarity metric \cite{SimilarityMetric} is a Kolmogorov complexity based measure and it is computable for example with the use of CompLearn \cite{CompLearn}). Another possibility is to investigate sources as a directed graph of generalized function calls\footnote{For example, as like in C++ a+b; $\to$ operator+(a, b); and in a similar way it can be imagine that for(stat1; expr1; expr2) stat2; $\to$ statementfor(stat1, expr1, expr2, stat2); and so on.}. In this case, the generalized functions can be sorted by their PageRank \cite{PR} values. But this order can make a recursive statement: a function that is called by better functions is better\footnote{We had made similar measurements with running codes using AspectJ in paper \cite{OOmothetlang}.}.

How can we measure the development of computer programs? As it is well known, the Kolmogorov complexity has a strong relationship with the entropy \cite[pp. 69]{Lovasz}. Consider the fictional computing device shown in figure \ref{fig_fboxs}. In these boxes, the programming is implemented by the internal arrangements of walls. The input particles are prepared on the left side and the output will appear on the right side as the time goes on\footnote{Because of the probability nature of this computing device, we cannot give an accurate upper bound of its execution time.}. Using the notation of figure \ref{fig_fboxs}, $K(x \mapsto rnd(x)) > K(x \mapsto x) \simeq K(x \mapsto 2x)$
intuitively shows that the entropy changes in the same direction as the Kolmogorov complexity. But these and similar imaginable measures tell us nothing about the questions: does the code meet its requirements or does the code fit for a particular purpose. This is why the test-driven development methodology uses more or less independent tests in today's software engineering practices.

\begin{figure}[t!]
	\centering\scalebox{.8}{\input{nnk}}\caption{This figure shows a neural network between two consecutive training steps.\label{fig_nnk}}
	\vspace{-0.5cm}
\end{figure}
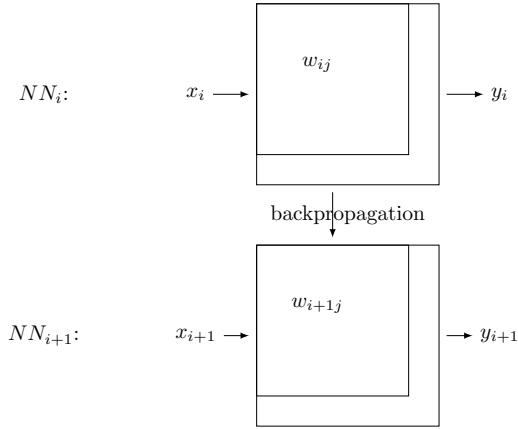

In contrast, in the neural computing paradigm the evaluation of the fitness of the programs for a particular purpose (aka the artificial neural networks) is an essential feature. Consider the training and testing of the networks, they are investigated roughly in the same fashion. For example, in a classification task, if we use TensorFlow, we will typically examine accuracy curves in TensorBoard. Consider a practical question: after 10 training steps, is the complexity of a neural network the same as after 1000 steps? Based on figure \ref{fig_nnk}, it is clearly observed that $K(NN_{i+1}) \le \vert x_i\vert + \vert NN_i\vert$. If we suppose that the initial weight vector $w_{0j}$ is selected from a random distribution, then the equation suggests that the complexity (together with the entropy) is the biggest at the beginning. Both this and the previous figure contradict with our naive intuition that programming increases complexity. Therefore, in GTB, we must try to measure the sophistication level of the program GTB based on its fitness for a particular purpose. We will try to convert the usually applied accuracy into a measure that could be computed in bits. Our vision is to create the roots of a new type of gamification by using some similar measure that allows casual gamers to participate in the revolution of artificial intelligence. The idea is to assign the measured values to the elements of the interface in GTB. The first suggested theme for the game interface is the Motherboard Builder introduced in section \ref{motherboard}. An imaginary element that has already been closely integrated with this theme can be seen in figure \ref{fig_procs}. This figure shows an abstraction of a processor that has 2.6 bits of information accuracy.

%% file: fbox1.tex
\begin{tikzpicture}

\draw  (-2.5,3) rectangle (2,-1.5);
\draw  (-3.5,3) rectangle (-2.5,-1.5);
\node (v2) at (-4.5,.75) {};
\node (v3) at (-3.5,0.75) {};
\draw [-latex] (v2) edge (v3);
\node (v4) at (3,.75) {};
\node (v5) at (4,.75) {};
\draw [-latex] (v4) edge (v5);
\node at (-4,1) {in};
\node at (3.5,1) {out};
\node at (-3,3.25) {x};
\node at (2.5,3.25) {y};

\draw  (2,3) rectangle (3,-1.5);
\node (v7) at (-2.5,0.75) {};

\node (v1) at (-3.5,1.8) {};
\node (v6) at (-2.5,1.8) {};

\node (v8) at (-3.5,-0.35) {};
\node (v9) at (-2.5,-0.35) {};

\node (v12) at (2,0.75) {};
\node (v10) at (2,1.8) {};
\node (v13) at (2,-0.4) {};
\node (v14) at (3,-0.4) {};
\node (v11) at (3,1.8) {};
\draw  (v1) edge (v6);
\draw  (v3) edge (v7);
\draw  (v8) edge (v9);
\draw  (v10) edge (v11);
\draw  (v4) edge (v12);
\draw  (v13) edge (v14);
\draw  (-3,2.5) ellipse (0.2 and 0.2);

\draw  (-3,.5) ellipse (0.2 and 0.2);
\draw  (-3,-1) ellipse (0.2 and 0.2);

\draw [-latex][dashed](-3,-1) -- (-1.5,0) -- (-1,-0.5) -- (-1,0) -- (-1.5,1.5) -- (0,1.5) -- (-0.5,0) -- (1,1) -- (-0.5,2.5) -- (1.5,2) -- (1,-0.5) -- (2.5,-1.2);

\draw  (2.5,-1.2) ellipse (0.2 and 0.2);
\draw  (2.6,-.7) ellipse (0.2 and 0.2);

\draw  (2.5,1.2) ellipse (0.2 and 0.2);

\node at (-4,0.5) {1011};
\node at (3.5,0.5) {0101};
\end{tikzpicture}

%% file: fbox2.tex
\begin{tikzpicture}

\draw  (-2.5,3) rectangle (2,-1.5);
\draw  (-3.5,3) rectangle (-2.5,-1.5);
\node (v2) at (-4.5,.75) {};
\node (v3) at (-3.5,0.75) {};
\draw [-latex] (v2) edge (v3);
\node (v4) at (3,.75) {};
\node (v5) at (4,.75) {};
\draw [-latex] (v4) edge (v5);
\node at (-4,1) {in};
\node at (3.5,1) {out};
\node at (-3,3.25) {x};
\node at (2.5,3.25) {y};

\draw  (2,3) rectangle (3,-1.5);
\node (v7) at (-2.5,0.75) {};

\node (v1) at (-3.5,1.8) {};
\node (v6) at (-2.5,1.8) {};

\node (v8) at (-3.5,-0.35) {};
\node (v9) at (-2.5,-0.35) {};

\node (v12) at (2,0.75) {};
\node (v10) at (2,1.8) {};
\node (v13) at (2,-0.4) {};
\node (v14) at (3,-0.4) {};
\node (v11) at (3,1.8) {};
\draw  (v1) edge (v6);
\draw  (v3) edge (v7);
\draw  (v8) edge (v9);
\draw  (v10) edge (v11);
\draw  (v4) edge (v12);
\draw  (v13) edge (v14);
\draw  (-3,2.5) ellipse (0.2 and 0.2);

\draw  (-3,.5) ellipse (0.2 and 0.2);
\draw  (-3,-1) ellipse (0.2 and 0.2);

\draw  (2.5,-1.2) ellipse (0.2 and 0.2);
\draw  (2.6,0) ellipse (0.2 and 0.2);

\draw  (2.5,2.2) ellipse (0.2 and 0.2);

\node at (-4,0.5) {1011};
\node at (3.5,0.5) {1011};
\draw  (v6) edge (v10);
\draw  (v7) edge (v12);
\draw  (v9) edge (v13);
\draw [-latex][dashed] (-3,-1) -- (-1.5,-0.5) -- (-2,-1.5) -- (-1,-1) -- (-1,-0.5) -- (-0.5,-1.5) -- (0,-0.5) -- (1.5,-0.5) -- (0.5,-1) -- (2.5,-1.2);

\end{tikzpicture}

%% file: fbox3.tex
\begin{tikzpicture}

\draw  (-2.5,3) rectangle (2,-1.5);
\draw  (-3.5,3) rectangle (-2.5,-1.5) node (v16) {};
\node (v2) at (-4.5,.75) {};
\node (v3) at (-3.5,0.75) {};
\draw [-latex] (v2) edge (v3);
\node (v4) at (3,.75) {};
\node (v5) at (4,.75) {};
\draw [-latex] (v4) edge (v5);
\node at (-4,1) {in};
\node at (3.5,1) {out};
\node at (-3,3.25) {x};
\node at (2.5,3.25) {y};

\draw  (2,3) node (v15) {} rectangle (3,-1.5);
\node (v7) at (-2.5,0.75) {};

\node (v1) at (-3.5,1.8) {};
\node (v6) at (-2.5,1.8) {};

\node (v8) at (-3.5,-0.35) {};
\node (v9) at (-2.5,-0.35) {};

\node (v12) at (2,0.75) {};
\node (v10) at (2,1.8) {};
\node (v13) at (2,-0.4) {};
\node (v14) at (3,-0.4) {};
\node (v11) at (3,1.8) {};
\draw  (v1) edge (v6);
\draw  (v3) edge (v7);
\draw  (v8) edge (v9);
\draw  (v10) edge (v11);
\draw  (v4) edge (v12);
\draw  (v13) edge (v14);
\draw  (-3,2.5) ellipse (0.2 and 0.2);

\draw  (-3,.5) ellipse (0.2 and 0.2);
\draw  (-3,-1) ellipse (0.2 and 0.2);
\draw  (2.6,0) ellipse (0.2 and 0.2);

\draw  (2.5,1.2) node (v17) {} ellipse (0.2 and 0.2);

\node at (-4,0.5) {1011};
\node at (3.5,0.5) {0110};
\draw  (v6) edge (v15);
\draw  (v7) edge (v10);
\draw  (v9) edge (v12);
\draw  (v16) edge (v13);
\draw [-latex][dashed] (-3,0.5) -- (-2,0) -- (-2,0.5) -- (-0.5,1) -- (-1,0) -- (-0.5,0.5) -- (1,1) -- (0.5,1) -- (1.5,1.5) -- (1.5,1) -- (v17);
\end{tikzpicture}

%% file: nnk.tex
\begin{tikzpicture}

\draw  (-1,2.5) node (v1) {} rectangle (2,-0.5);
\draw  (v1) rectangle (1.5,0);
\node (v2) at (-2,1) {$x_i$};
\node (v3) at (-1,1) {};
\node (v4) at (2,1) {};
\node (v5) at (3,1) {$y_i$};
\node (v6) at (0.25,-0.5) {};
\node  (v7) at (0.25,-1.5) {};
\draw [-latex] (v2) edge (v3);
\draw [-latex] (v4) edge (v5);
\draw [-latex] (v6) edge (v7);
\draw  (-1,-1.5) node (v8) {} rectangle (2,-4.5);
\draw  (v8) rectangle (1.5,-4);
\node (v9) at (-2,-3) {$x_{i+1}$};
\node (v10) at (-1,-3) {};
\node (v12) at (2,-3) {};
\node (v11) at (3,-3) {$y_{i+1}$};
\draw [-latex] (v9) edge (v10);
\draw [-latex] (v12) edge (v11);
\node at (0,1.5) {$w_{ij}$};
\node at (0,-2.5) {$w_{i+1j}$};
\node at (-4.5,1) {$NN_i$:};
\node at (-4.5,-3) {$NN_{i+1}$:};
\node at (0.5,-1) {backpropagation};
\end{tikzpicture}

%% file: infacc.tex
\subsubsection*{Information Accuracy} At this point, we present an example to illustrate how the usual accuracy can be converted into information. Let us consider the following example\footnote{It can be seen in detail in the forked TensorFlow repository \url{https://github.com/nbatfai/tensorflow/blob/master/tensorflow/examples/tutorials/mnist}. The presented Python code snippet can be found in the file called \texttt{mnist\_softmax\_esmu.py}.}.

Let 
$Y\!\verb|_|=\left(y\!\verb|_|_i\right)$
denote the sequence of labels of input images where $y\!\verb|_|_i \in \{0, 1\}^n$ is a one-hot vector corresponded to an input image\footnote{In this example $n$ equals 10 due to a label is a digit between 0 and 9.}. And let 
$Y=\left(p_i\right)$
denote the probabilities of classifications where $p_i \in \mathbb{R}^n$ such that $\sum_{j=1}^{n}p_i(j) = 1$.
Then, using the entropies of labels $E=\left(e_i\right)$, $e_i \in \mathbb{R}$,  $e_i =-\sum_{j=1}^{n}p_i(j) log_2p_i(j)$
we can define the information accuracy as follow
$infoacc = \sum_{i=1}^{m} a_i e_i$ where 
\[ a_i=\left\{
\begin{array}{ll}
-1 & \textrm{if }
y_i \ne y\!\verb|_|_i
\\
1&
\textrm{otherwise}
\end{array}
\right.
\] that is $a_i$ simply shows that the classification of the i-th image is good or bad and $m$ is the number of classified images in the sequence $Y\!\verb|_|$. Or to be more  readable and precise

\begin{lstlisting}[basicstyle=\footnotesize, language=Python]
  plogp = tf.multiply(prob, tf.log(prob)/tf.log(2.0))
  plogpsum = -tf.reduce_sum(plogp, 1, keep_dims=True)
  plusminus = tf.cast(correct_prediction, 
	tf.float32)*2.0-1.0
  plusminus = tf.reshape(plusminus, [batchsize, 1])
  info = tf.multiply(plogpsum, plusminus)
  infoacc = tf.reduce_sum(info) / 
	tf.cast(batchsize, tf.float32)
\end{lstlisting}
where the tensor $prob$ holds the vectors of probabilities of classifications and $batchsize$ is the number of images.

\begin{figure}[ht!]
\centering
\begin{subfigure}[t]{0.4\textwidth}
	\centering\scalebox{.5}{\input{proc}}\caption{This figure shows the typical use cases of a neural network based program where we replace the usual accuracy by information accuracy.\label{fig_proc}}
\end{subfigure}
\quad
\begin{subfigure}[t]{0.4\textwidth}
	\centering\includegraphics[scale=.25]{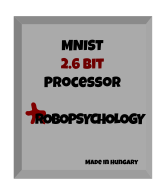}\caption{This is an imaginary chip which represents the gamification of the MNIST graph of the TensorFlow tutorial.\label{fig_mnistproc}}
\end{subfigure}
\caption{A gamification of a whole network is shown in general in \ref{fig_proc} and a specific gamification element is shown in \ref{fig_mnistproc}. The label ROBOPSYCHOLOGY on the chip is only a design element that suggests that there is a higher interpretation level of our work. See \cite{RoboPsyGitHub} and \cite{RoboPsyManifesto} for more detail.\label{fig_procs}}
\end{figure}

\begin{figure}
	\centering\includegraphics[scale=.35]{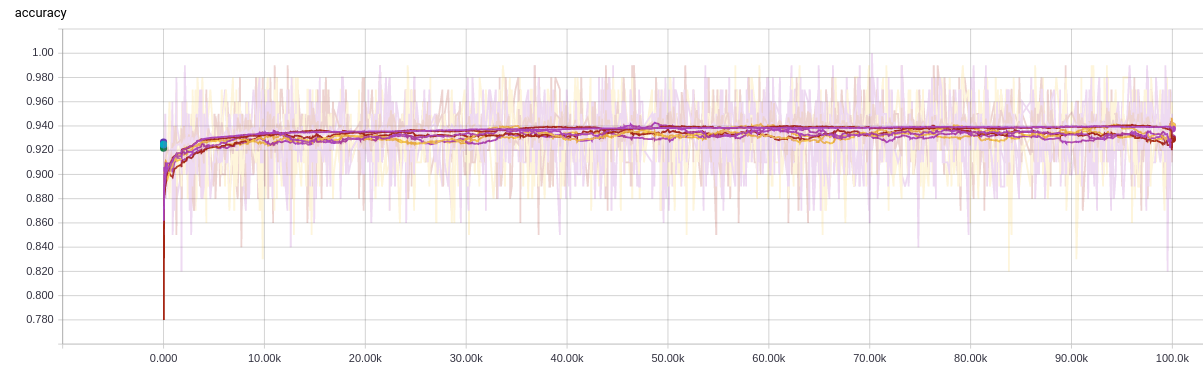}
	\caption{This figure shows the usual accuracy curves of MNIST tutorial in TensorBoard.\label{fig_acc}}
\end{figure}

\begin{figure}
	\centering\includegraphics[scale=.35]{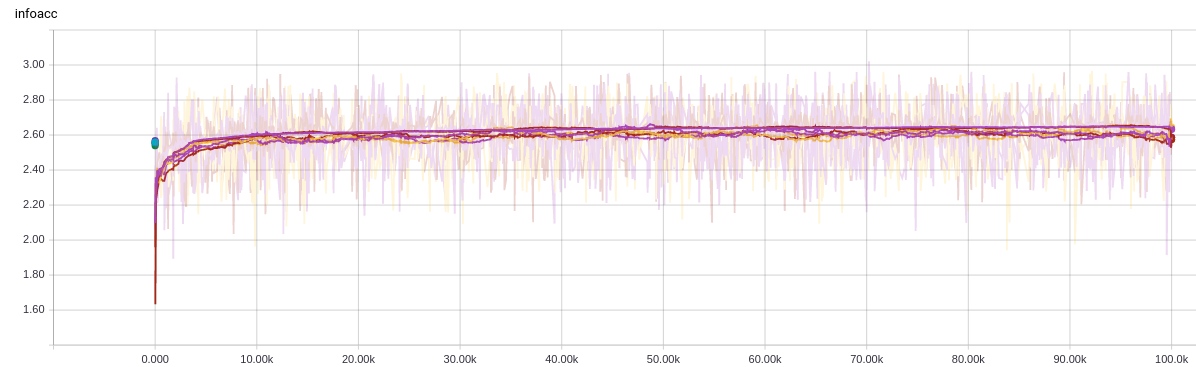}
	\caption{This figure shows the information accuracy curves corresponding to the curves in figure \ref{fig_acc}.\label{fig_infoacc}}
\end{figure}

Figure \ref{fig_acc} shows some usual accuracy curves with $batchsize$ of $10$, $100$, $1000$ and $10000$. The corresponding information accuracy curves are shown in figure \ref{fig_infoacc}. We can observe that the shapes of the curves are the same, but the scale has changed. So, the \enquote{learning} of the neural network architecture is followed by the information accuracy of the learning procedure.

%% file: proc.tex
\begin{tikzpicture}

\draw  (-2.5,3) rectangle (2,-1.5);
\draw  (-2.5,4) rectangle (2,3) node (v1) {};
\draw  (-2.5,-1.5) rectangle (2,-2.5);
\draw  (-3.5,3) rectangle (-2.5,-1.5);
\draw  (v1) rectangle (3,-1.5);
\node (v2) at (-4.5,.75) {};
\node (v3) at (-3.5,0.75) {};
\draw [-latex] (v2) edge (v3);
\node (v4) at (3,.75) {};
\node (v5) at (4,.75) {};
\draw [-latex] (v4) edge (v5);
\node (v6) at (-0.5,5) {};
\node (v7)at (-0.5,4) {};
\draw [-latex] (v6) edge (v7);
\node (v8) at (-0.25,-2.5) {};
\node (v9) at (-0.25,-3.5) {};
\draw [-latex] (v8) edge (v9);
\node at (-0.5,3.5) {$\{(x,y)\}$};
\node at (-4,1) {in};
\node at (3.5,1) {out};
\node at (-3,0.75) {x};
\node at (2.5,0.75) {y};
\node at (-0.5,5.5) {train/test};
\node at (-0.25,-2) {infoacc};
\end{tikzpicture}

%% file: conclusion.tex
\section{Conclusion}

From a pure theoretical point of view, our present work can be regarded as a set of preliminary studies regarding the following questions: What notion of complexity levels are we looking for to the gamification of computing graphs? How can we measure the aliveness of the code? How could the code resist against increasing entropy? The latter two questions have already focused on the approach introduced by Erwin Schr\"odinger in his famous book \cite{WhatIsLife} to investigate living systems. To apply his ideas in the field of computer science seems a very interesting challenge\footnote{In the terminology of Turing machines, we have already formulated a similar question in \url{https://github.com/nbatfai/AlgorithmicFractals/tree/master/manuscript}.}. The thought experiments shown in figure \ref{fig_fboxs} and \ref{fig_nnk} contradicts our intuitive expectation that programming increases complexity (entropy), therefore in this sense, programming is similar to entropy non-increasing behavior of living systems.

The work described in this paper has formed and crystallized our main research question: how can a TensorFlow computational graph be visualized on the surface of a massively multiplayer game? Simplifying a TF graph without losing important components and providing a possibility to fine-tune it by non-expert gamers at the same time seems an interesting challenge. We think, answering this question is crucial to making a big leap towards the manifestation of the conception of widely open AI. Google with the TensorFlow API have simplified the prototyping of ML architectures and have made it easier to experiment with new ideas, even though it requires broad knowledge in AI methods and algorithms. In our vision, a powerful tool that can speed up the evolution of ML by widening the group of people who working with it, can be applications that implement the ESAMU concept. By playing these games, which we consider as an embryonic form of robopsychology, we may get closer to the creation of The Artificial Intelligence in its universal form.

%% file: ack.tex
\section*{Acknowledgment}

The authors would like to thank the students of the BSc course of \enquote{High Level Programming Languages} in the fall semester of 2016/2017 at the University of Debrecen and the community members of the group UDPROG\footnote{\url{https://www.facebook.com/groups/udprog/}} for their interest and for their testing and contribution to the repository of ESAMU \cite{ESAMUGitHub}. Special thanks to the members of the mailing list desport\footnote{\url{https://groups.google.com/forum/#!forum/desport-desamu}}. The authors also would like to thank Mih\'aly Szil\'agyi and Louis Joseph Mattia for the proofreading of the paper.

%% file: faba.bbl
\newcommand{\etalchar}[1]{$^{#1}$}
\begin{thebibliography}{LMHR{\etalchar{+}}16}

\bibitem[AAB{\etalchar{+}}16]{TF}
Mart{\'{\i}}n Abadi, Ashish Agarwal, Paul Barham, Eugene Brevdo, Zhifeng Chen,
  Craig Citro, Gregory~S. Corrado, Andy Davis, Jeffrey Dean, Matthieu Devin,
  Sanjay Ghemawat, Ian~J. Goodfellow, Andrew Harp, Geoffrey Irving, Michael
  Isard, Yangqing Jia, Rafal J{\'{o}}zefowicz, Lukasz Kaiser, Manjunath Kudlur,
  Josh Levenberg, Dan Man{\'{e}}, Rajat Monga, Sherry Moore, Derek~Gordon
  Murray, Chris Olah, Mike Schuster, Jonathon Shlens, Benoit Steiner, Ilya
  Sutskever, Kunal Talwar, Paul~A. Tucker, Vincent Vanhoucke, Vijay Vasudevan,
  Fernanda~B. Vi{\'{e}}gas, Oriol Vinyals, Pete Warden, Martin Wattenberg,
  Martin Wicke, Yuan Yu, and Xiaoqiang Zheng.
\newblock {TensorFlow}: Large-scale machine learning on heterogeneous
  distributed systems.
\newblock {\em CoRR}, abs/1603.04467, 2016.

\bibitem[Aut17]{TFGitHub}
The~TensorFlow Authors.
\newblock Computation using data flow graphs for scalable machine learning
  http://tensorflow.org.
\newblock \url{https://github.com/tensorflow/tensorflow}, 2017.
\newblock Online; accessed 22 January 2017.

\bibitem[B{\'a}t11]{OOmothetlang}
Norbert B{\'a}tfai.
\newblock Do the object oriented programs have a mother tongue: or an
  introduction of an analytical weaving (\textit{Van-e az objektum-orient\'alt
  programoknak anyanyelve? – avagy egy analitikai sz\"ov\'es bevezet\'ese}).
\newblock {\em H\'irad\'astechnika}, LXVI(2):27--32, 2011.
\newblock Original paper in Hungarian.

\bibitem[B{\'a}t17a]{RoboPsyGitHub}
Norbert B{\'a}tfai.
\newblock {Robopsychology}.
\newblock
  \url{https://github.com/nbatfai/Robopsychology/files/169195/robopsychology.pdf},
  2017.
\newblock Online; accessed 2 February 2017.

\bibitem[B{\'a}t17b]{ESAMUGitHub}
Norbert B{\'a}tfai.
\newblock {Samu Entropy: ESPORT AND ARTIFICIAL INTELLIGENCE}.
\newblock \url{https://github.com/nbatfai/SamuEntropy}, 2017.
\newblock Online; accessed 6 January 2017.

\bibitem[BB17]{RoboPsyManifesto}
Norbert B{\'a}tfai and Ren{\'a}t{\'o} Besenczi.
\newblock {Robopsychology manifesto: Samu in his prenatal development}.
\newblock {\em submitted manuscript}, 2017.

\bibitem[BBL{\etalchar{+}}17]{ESAMUInfTars}
Norbert B{\'a}tfai, M{\'a}ri{\'o} Bersenszki, Mikl{\'o}s Luk{\'a}cs,
  Ren{\'a}t{\'o} Besenczi, Gerg{\H o} Bogacsovics, and P{\'e}ter Jeszenszky.
\newblock {The common future of e-sport and robopsychology (\textit{Az e-sport
  {\'e}s a robotpszichol{\'o}gia k{\"o}z{\"o}s j{\"o}v{\H o}je})}.
\newblock {\em submitted manuscript}, 2017.
\newblock Original paper in Hungarian.

\bibitem[BBS16]{ESAMUDevGuide}
N.~B\'atfai, R.~Besenczi, and V.~Simk\'o.
\newblock {\em Samu Entropy developer's guide}.
\newblock \url{https://github.com/nbatfai/SamuEntropy/tree/master/docs}, 2016.
\newblock translators: A. Fodor, G. Bogacsovics, J. Pog\'any.

\bibitem[BGNR16]{CNNArc}
Bowen Baker, Otkrist Gupta, Nikhil Naik, and Ramesh Raskar.
\newblock Designing neural network architectures using reinforcement learning.
\newblock {\em CoRR}, abs/1611.02167, 2016.

\bibitem[CTB{\etalchar{+}}10]{Foldit}
Seth Cooper, Adrien Treuille, Janos Barbero, Andrew Leaver-Fay, Kathleen Tuite,
  Firas Khatib, Alex~Cho Snyder, Michael Beenen, David Salesin, David Baker,
  and Zoran Popovi\'{c}.
\newblock {The challenge of designing scientific discovery games}.
\newblock In {\em {Proceedings of the Fifth International Conference on the
  Foundations of Digital Games}}, {FDG '10}, pages 40--47. ACM, 2010.

\bibitem[CV05]{CompLearn}
Rudi Cilibrasi and Paul M.~B. Vit{\'a}nyi.
\newblock Clustering by compression.
\newblock {\em IEEE Transactions on Information Theory}, 51:1523--1545, 2005.

\bibitem[Gol16]{TFTour}
Peter Goldsborough.
\newblock {A tour of TensorFlow}.
\newblock {\em CoRR}, abs/1610.01178, 2016.

\bibitem[Kah74]{Kahn}
G.~Kahn.
\newblock The semantics of a simple language for parallel programming.
\newblock In {\em Information processing}, pages 471--475. North Holland,
  Amsterdam, 1974.

\bibitem[KSH12]{ImageNet}
Alex Krizhevsky, Ilya Sutskever, and Geoffrey~E Hinton.
\newblock {ImageNet} classification with deep convolutional neural networks.
\newblock In F.~Pereira, C.~J.~C. Burges, L.~Bottou, and K.~Q. Weinberger,
  editors, {\em Advances in Neural Information Processing Systems 25}, pages
  1097--1105. Curran Associates, Inc., 2012.

\bibitem[KSSMB16]{7780896}
I.~Kemelmacher-Shlizerman, S.~M. Seitz, D.~Miller, and E.~Brossard.
\newblock The megaface benchmark: 1 million faces for recognition at scale.
\newblock In {\em IEEE Conference on Computer Vision and Pattern Recognition
  (CVPR)}, pages 4873--4882, 2016.

\bibitem[LBBH98]{LeCun98}
Y.~LeCun, L.~Bottou, Y.~Bengio, and P.~Haffner.
\newblock Gradient-based learning applied to document recognition.
\newblock {\em {Proceedings of the IEEE}}, 86(11):2278--2324, 1998.

\bibitem[LCL{\etalchar{+}}04]{SimilarityMetric}
Ming Li, Xin Chen, Xin Li, Bin Ma, and P.M.B. Vit{\'a}nyi.
\newblock The similarity metric.
\newblock {\em IEEE Transactions on Information Theory}, 50(12):3250--3264,
  2004.

\bibitem[LLWT15]{liu2015faceattributes}
Z.~Liu, P.~Luo, X.~Wang, and X.~Tang.
\newblock Deep learning face attributes in the wild.
\newblock In {\em IEEE International Conference on Computer Vision (ICCV)},
  pages 3730--3738, 2015.

\bibitem[LMHR{\etalchar{+}}16]{LFW}
Erik Learned-Miller, Gary~B. Huang, Aruni RoyChowdhury, Haoxiang Li, and Gang
  Hua.
\newblock Labeled faces in the wild: A survey.
\newblock In Michal Kawulok, M.~Emre Celebi, and Bogdan Smolka, editors, {\em
  Advances in Face Detection and Facial Image Analysis}, pages 189--248.
  Springer International Publishing, 2016.

\bibitem[Lov94]{Lovasz}
L{\'a}szl{\'o} Lov{\'a}sz.
\newblock {\em {Algoritmusok bonyolults{\'a}ga}}.
\newblock Nemzeti Tank{\"o}nyvkiad{\'o}, 1994.

\bibitem[{McC}76]{CyclomaticComplexity}
{McCabe, T. J.}
\newblock A complexity measure.
\newblock {\em IEEE Trans. Softw. Eng.}, 2(4):308--320, 1976.

\bibitem[MKS{\etalchar{+}}15]{DMNature1}
V~Mnih, K~Kavukcuoglu, D~Silver, AA~Rusu, J~Veness, MG~Bellemare, A~Graves,
  M~Riedmiller, AK~Fidjeland, G~Ostrovski, S~Petersen, C~Beattie, A~Sadik,
  I~Antonoglou, H~King, D~Kumaran, D~Wierstra, S~Legg, and D.~Hassabis.
\newblock Human-level control through deep reinforcement learning.
\newblock {\em Nature}, 518:529--533, 2015.

\bibitem[NW14]{7025068}
H.~W. Ng and S.~Winkler.
\newblock A data-driven approach to cleaning large face datasets.
\newblock In {\em IEEE International Conference on Image Processing (ICIP)},
  pages 343--347, 2014.

\bibitem[PBMW99]{PR}
Lawrence Page, Sergey Brin, Rajeev Motwani, and Terry Winograd.
\newblock {The PageRank citation ranking: bringing order to the web.}
\newblock Technical Report 1999-66, Stanford InfoLab, 1999.

\bibitem[Pen16]{PenroseEmperor}
Roger Penrose.
\newblock {\em The Emperor's new mind: Concerning computers, minds, and the
  laws of physics}.
\newblock Oxford University Press, 2016.

\bibitem[PHS15]{csgames}
Marisa Ponti, Thomas Hillman, and Igor Stankovic.
\newblock Science and gamification: The odd couple?
\newblock In {\em Proceedings of the 2015 Annual Symposium on Computer-Human
  Interaction in Play}, CHI PLAY '15, pages 679--684. ACM, 2015.

\bibitem[Sch44]{WhatIsLife}
Erwin Schr{\"o}dinger.
\newblock {\em {What is life?: the physical aspect of the living cell}}.
\newblock Cambridge University Press, 1944.

\bibitem[SHM{\etalchar{+}}16]{DMNature2}
D~Silver, A~Huang, CJ~Maddison, A~Guez, L~Sifre, G~van~den Driessche,
  J~Schrittwieser, I~Antonoglou, V~Panneershelvam, M~Lanctot, S~Dieleman,
  D~Grewe, J~Nham, N~Kalchbrenner, I~Sutskever, T~Lillicrap, M~Leach,
  K~Kavukcuoglu, T~Graepel, and D.~Hassabis.
\newblock Mastering the game of {Go} with deep neural networks and tree search.
\newblock {\em Nature}, 529:484--503, 2016.

\bibitem[SKP15]{FaceNet}
Florian Schroff, Dmitry Kalenichenko, and James Philbin.
\newblock {FaceNet}: A unified embedding for face recognition and clustering.
\newblock {\em CoRR}, abs/1503.03832, 2015.

\bibitem[SLJ{\etalchar{+}}14]{GoogLeNet}
Christian Szegedy, Wei Liu, Yangqing Jia, Pierre Sermanet, Scott~E. Reed,
  Dragomir Anguelov, Dumitru Erhan, Vincent Vanhoucke, and Andrew Rabinovich.
\newblock Going deeper with convolutions.
\newblock {\em CoRR}, abs/1409.4842, 2014.

\bibitem[Sou12]{DFP}
T.~B. Sousa.
\newblock Dataflow programming concepts, languages and applications.
\newblock \url{http://paginas.fe.up.pt/prodei/dsie12/papers/paper_17.pdf},
  2012.
\newblock Online; accessed 12 January 2017.

\bibitem[SVI{\etalchar{+}}15]{Inception3}
Christian Szegedy, Vincent Vanhoucke, Sergey Ioffe, Jonathon Shlens, and
  Zbigniew Wojna.
\newblock {Rethinking the Inception Architecture for Computer Vision}.
\newblock {\em CoRR}, abs/1512.00567, 2015.

\bibitem[Swa09]{SGU}
James Swallow.
\newblock {\em {SGU Stargate Universe}}.
\newblock Fandemonium Books, New York City, 2009.

\bibitem[TSBW16]{hcgames}
Olivier Tremblay-Savard, Alexander Butyaev, and J{\'e}r\^{o}me Waldisp\"{u}hl.
\newblock Collaborative solving in a human computing game using a market,
  skills and challenges.
\newblock In {\em Proceedings of the 2016 Annual Symposium on Computer-Human
  Interaction in Play}, CHI PLAY '16, pages 130--141. ACM, 2016.

\bibitem[TYRW14]{DeepFace}
Yaniv Taigman, Ming Yang, Marc'Aurelio Ranzato, and Lior Wolf.
\newblock {DeepFace}: Closing the gap to human-level performance in face
  verification.
\newblock In {\em {Proceedings of the 2014 IEEE Conference on Computer Vision
  and Pattern Recognition}}, CVPR '14, pages 1701--1708, 2014.

\bibitem[VSR{\etalchar{+}}16]{EvolArc}
Arso~M Vukicevic, Miroslav Stojadinovic, Milos Radovic, Milena Djordjevic,
  Bojana~Andjelkovic Cirkovic, Tomislav Pejovic, Gordana Jovicic, and Nenad
  Filipovic.
\newblock Automated development of artificial neural networks for clinical
  purposes: Application for predicting the outcome of choledocholithiasis
  surgery.
\newblock {\em Computers in Biology and Medicine}, 75:80--89, 2016.

\bibitem[YLLL14]{yi2014learning}
Dong Yi, Zhen Lei, Shengcai Liao, and Stan~Z Li.
\newblock Learning face representation from scratch.
\newblock {\em arXiv preprint arXiv:1411.7923}, 2014.

\bibitem[ZL16]{TFArc}
Barret Zoph and Quoc~V. Le.
\newblock Neural architecture search with reinforcement learning.
\newblock {\em CoRR}, abs/1611.01578, 2016.

\bibitem[ZZWS12]{6140979}
X.~Zhang, L.~Zhang, X.~J. Wang, and H.~Y. Shum.
\newblock Finding celebrities in billions of web images.
\newblock {\em IEEE Transactions on Multimedia}, 14(4):995--1007, 2012.

\end{thebibliography}
